%% file: main.tex
\documentclass[letterpaper, 10 pt, journal, twoside]{IEEEtran}

\IEEEoverridecommandlockouts

\usepackage{graphics} 
\usepackage{epsfig} 
\usepackage{times} 
\usepackage{amsmath} 
\usepackage{amssymb}  
\usepackage[table]{xcolor}
\usepackage[ruled,linesnumbered, noend]{algorithm2e}
\usepackage{flushend} 

\usepackage{soul}
\usepackage{cite}
\usepackage{comment}
\usepackage{multirow}
\usepackage{graphicx}
\usepackage{wrapfig}
\usepackage{soul, color}
\usepackage{wrapfig}
\usepackage{mathtools}
\usepackage{graphicx}
\usepackage{booktabs}
\usepackage{adjustbox}
\usepackage{siunitx}
\usepackage{hyperref}
\hypersetup{
    colorlinks=true,
    citecolor=black,
    linkcolor=black,
    filecolor=black,      
    urlcolor=blue,
    pdftitle={Overleaf Example},
    pdfpagemode=FullScreen,
    }

\usepackage{caption}
\usepackage{subfigure}

\usepackage{lipsum}

\definecolor{LightCyan}{rgb}{0.88,1,1}
\definecolor{Gray}{gray}{0.9}

\newcommand{\ve}[1]{\mathbf{#1}} 
\newcommand{\hve}[1]{\hat{\mathbf{#1}}} 

\title{Visuotactile-Based Learning for Insertion with Compliant Hands}
\author{Osher Azulay, Dhruv Metha Ramesh, Nimrod Curtis and Avishai Sintov
\thanks{O. Azulay, N. Curtis and A. Sintov are with the School of Mechanical Engineering, Tel-Aviv University, Israel. Corresponding Author: osherazulay@mail.tau.ac.il.}
\thanks{R.M. Ramesh is with the Computer Science Department of Rutgers University, New Brunswick, NJ, USA.}
\thanks{This work was supported by the Pazy Foundation (grant No. 283-20).}
}

\begin{document}

\setlength{\belowdisplayskip}{2pt}
\setlength{\belowdisplayshortskip}{3pt}
\setlength{\abovedisplayskip}{2pt} 
\setlength{\abovedisplayshortskip}{3pt}
\setlength{\parskip}{0pt}


\maketitle
\thispagestyle{empty}
\pagestyle{empty}


\input{abstract}

\section{Introduction}
\input{introduction}
\section{Related Work}
\label{sec:related}
\input{related}

\section{Method}

\input{method}

\section{Experiments}
\input{experiments}

\section{Conclusions}
\input{conclusions}

\bibliographystyle{IEEEtran}
\bibliography{ref}

\end{document}

%% file: abstract.tex
\begin{abstract}
Compared to rigid hands, underactuated compliant hands offer greater adaptability to object shapes, provide stable grasps, and are often more cost-effective. However, they introduce uncertainties in hand-object interactions due to their inherent compliance and lack of precise finger proprioception as in rigid hands. These limitations become particularly significant when performing contact-rich tasks like insertion. To address these challenges, additional sensing modalities are required to enable robust insertion capabilities. This letter explores the essential sensing requirements for successful insertion tasks with compliant hands, focusing on the role of visuotactile perception (i.e., visual and tactile perception). We propose a simulation-based multimodal policy learning framework that leverages all-around tactile sensing and an extrinsic depth camera. A transformer-based policy, trained through a teacher-student distillation process, is successfully transferred to a real-world robotic system without further training. Our results emphasize the crucial role of tactile sensing in conjunction with visual perception for accurate object-socket pose estimation, successful sim-to-real transfer and robust task execution. 

\textit{Project website}: \url{https://osheraz.github.io/visuotactile/}.
\end{abstract}

%% file: introduction.tex


Daily life activities inherently involve interacting with the environment through contact. For robots to perform real-world tasks effectively, they must navigate through unstructured, contact-rich environments. While humans execute various interaction tasks with relative ease, achieving similar levels of precision and reliability in such environments poses a significant challenge for robots \cite{zhao2023learning}. Physical interaction of the robot with its surroundings can greatly enhance perception and understanding of the environment \cite{azulay2022haptic}. These interactions provide essential information that is often otherwise unavailable, playing a critical role in adapting to uncertainties and improving robotic performance. This is particularly true for precision manipulation, where robots must perceive, grasp, manipulate, and accurately place objects according to task-specific requirements.

\begin{figure}
\centering
\includegraphics[width=\linewidth]{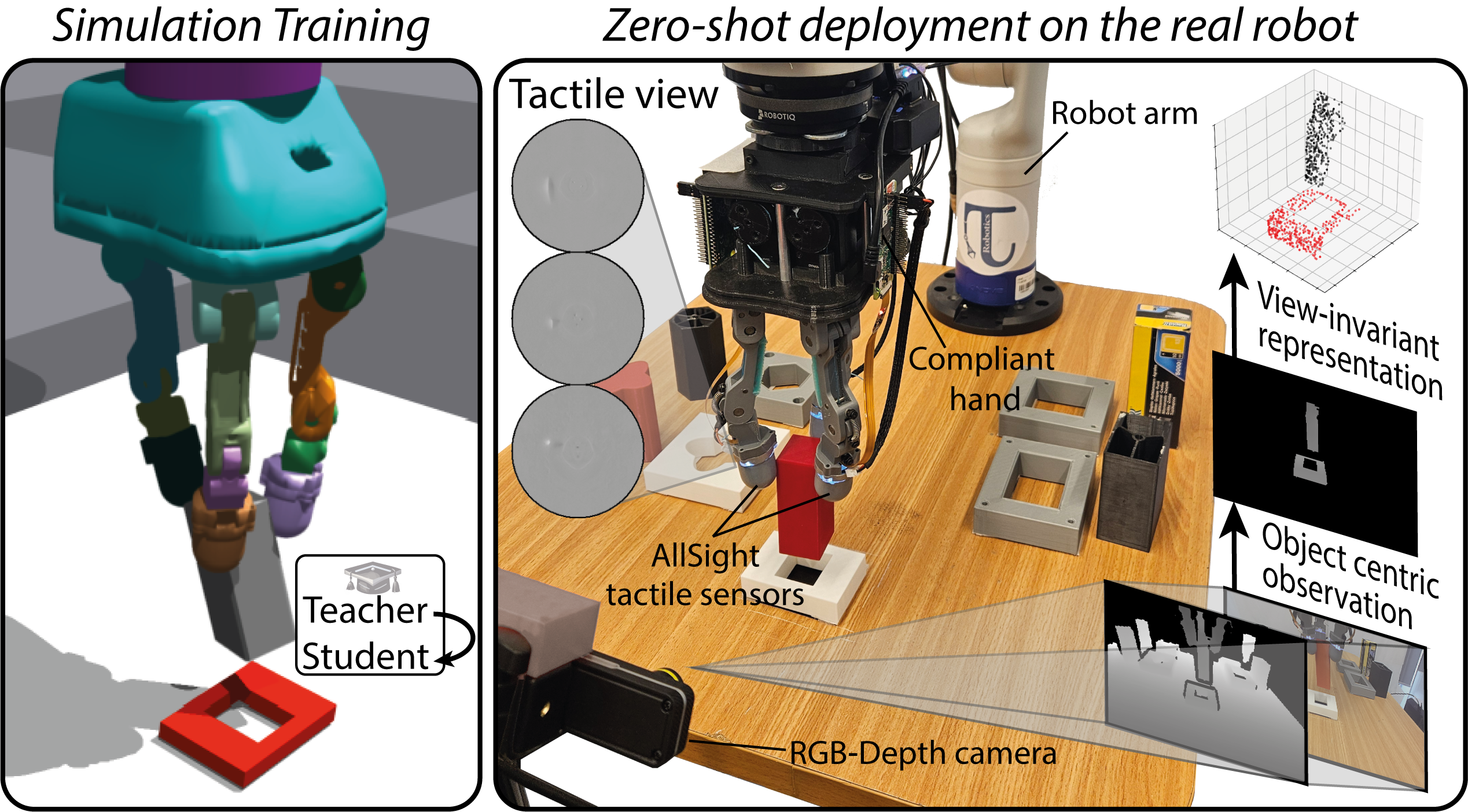} 
    \caption{\small Tight insertion of an object into a socket with a robotic arm and a three-finger compliant hand without hand proprioception. Two sensing modalities are used: vision provides a rough estimate of the object-socket poses and tactile sensors on the fingers deliver implicit contact information. A policy is initially trained in simulation and subsequently deployed in zero-shot on the real system.}
\label{fig:intro}
\vspace{-0.8cm}
\end{figure}

The majority of research in robot manipulation has focused on rigid hands. However, rigid hands can struggle to effectively handle uncertainties stemming from inaccurate object pose estimation \cite{Dollar2010}. On the other hand, compliant hands incorporate underactuated mechanisms, such as combinations of tendons and springs, to achieve flexibility and adaptability in grasping various objects without closed-loop control \cite{Odhner2011}. Unlike rigid hands, the true state of a compliant hand is usually not observable \cite{Belief2019}. That is, finger joint angles and loads cannot be extracted during grasping and manipulation, making real-world manipulation with the hand much more challenging. To bypass this challenge, we explore the use of multimodal perception in compliant hands for contact-rich manipulation skills. Recently, multimodal perception has been advanced for manipulation learning with rigid hands and usually includes tactile, visual, and proprioceptive sensing \cite{Hansen2022}.  The integration of tactile and visual information is often referred to as \textit{visuotactile} sensing. Visual perception provides a comprehensive view of the environment, aiding in the localization and orientation of objects while providing rough pose estimations; tactile perception allows robots to better understand relative and local information, to detect and respond to subtle changes in contact forces and object configuration, facilitating delicate manipulations that are challenging with vision alone \cite{suresh2023neural}; and proprioception is essential for the coordinated use of tactile and visual feedback \cite{qi2023general}. As mentioned, proprioception is usually not available in compliant hands.

In this letter, we investigate the essential requirements for multimodal perception in task-centric manipulation involving compliant hands. As a test case, we consider insertion tasks (Figure \ref{fig:intro}), which often serve as an ideal benchmark for evaluating the effectiveness of contact management in robotic manipulation \cite{azulay2022haptic,RoyoMiquel2023}. Insertion is common in applications like assembly \cite{thomas2018learning} and dense packing \cite{Wang2022}. Achieving successful tight object insertion with visuotactile sensing validates the robustness of the robotic system, demonstrating its ability to handle complex real-world scenarios that demand precise positioning and delicate manipulation. Furthermore, we assert that incorporating structured visuotactile representations into policy design is crucial for improving the ability to manage contact variations. Effective policy design for compliant hand manipulation requires object-centric, spatial-aware, and relative-sensitive visuotactile representations. These attributes enable focusing on task-relevant objects, reasoning in 3D space, and detecting subtle contact variations. To the best of the author’s knowledge, this is the first exploration of visuotactile sensing for compliant hands without relying on proprioceptive feedback. 

\begin{figure*}[t]
\centering
\includegraphics[width=0.9\textwidth]{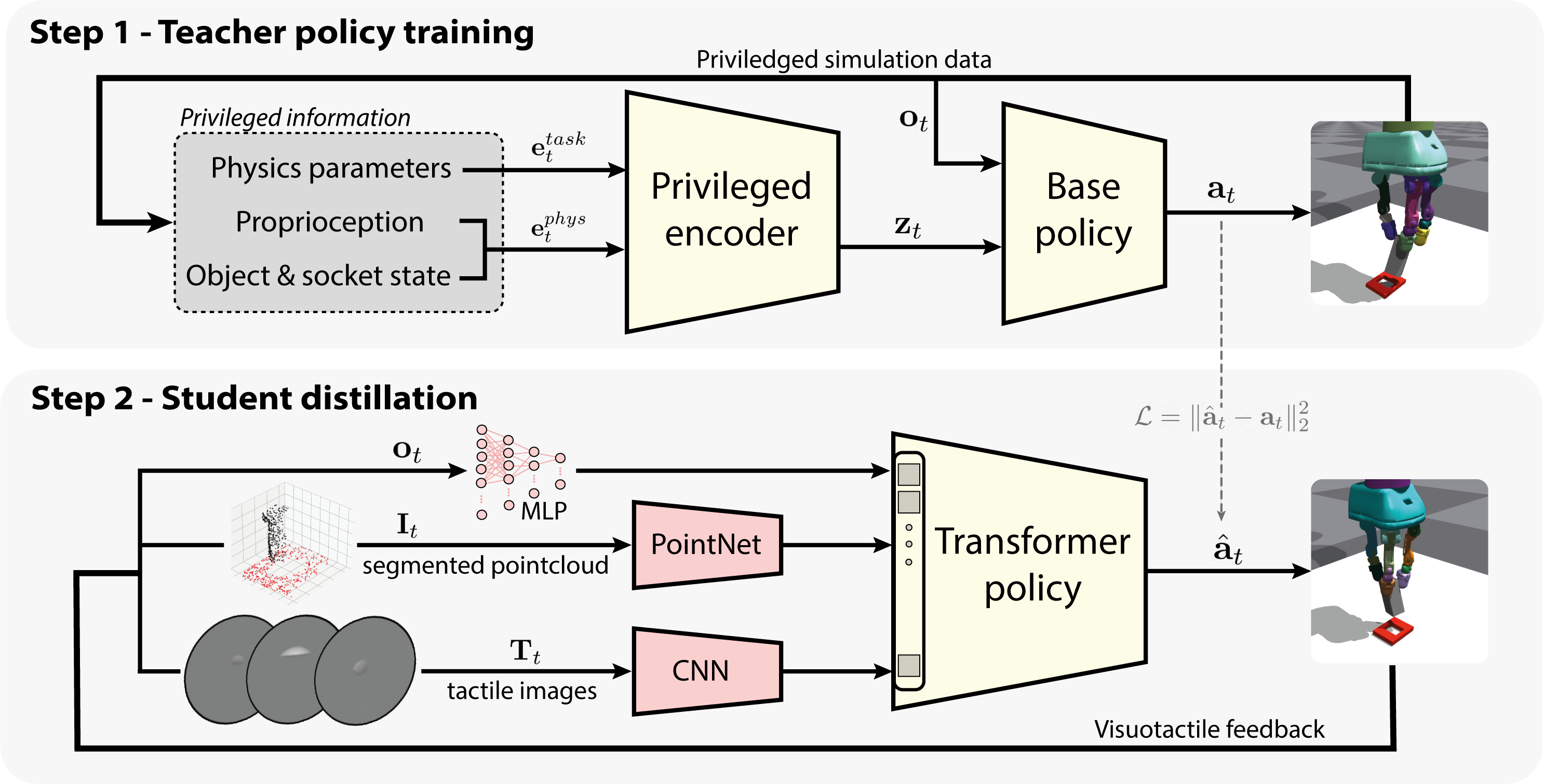} 
\caption{\small Overview illustration of the training steps in simulation. First, a teacher policy is trained using privileged information. Subsequently, a distillation process is employed to train a student policy that learns to imitate the teacher's behavior, relying solely on visuotactile data and End-Effector (EE) pose.}
\label{fig:sketch}
\vspace{-0.5cm}
\end{figure*}

In addition to understanding multimodal information for manipulation with compliant robotic hands, we explore simulative tools for ease of policy training with transfer to a real system. In particular, a simulation integrating visuotactile sensing and the compliant hand mechanism is used to train a teacher model with privileged information. Visuotactile sensing is composed of a masked point cloud observed by a depth camera, and the high-resolution AllSight \cite{azulay2023allsight} tactile sensor. Subsequently, an imitation learning approach is employed to distill the perceived data into its most essential components. This teacher-student learning is uniquely augmented by a probabilistic mechanism, enabling the student to learn robust behaviors while reducing over-reliance on the teacher policy. Then, the distilled policy is deployed on a real system with no further training. The policy is agnostic to the specific robotic arm used. Our methodology, illustrated in Figure \ref{fig:sketch}, highlights the importance of understanding multimodal information in enhancing the manipulation capabilities of compliant robotic hands. To conclude, the contributions of this letter are as follows:
\begin{enumerate}
    \item This work leverages compliant hands to address uncertain, contact-rich tasks with multimodal sensing.
    \item We investigate the novel use of multimodal perception for compliant robotic hands in contact-rich manipulation tasks, specifically focusing on insertion tasks.
    \item This work pioneers the use of visuotactile sensing for compliant hand manipulation without proprioception.
    \item We propose the integration of object-centric and spatial-aware visuotactile representations to manage contact variations in manipulation tasks.
    \item We use a simulation integrating visuotactile sensing and compliant hand mechanisms for policy training with zero-shot transfer to the real system.
    \item A Transformer policy uniquely handles visuotactile inputs for precise contact-rich manipulation.
    \item The simulation and deployment code are provided open-source\footnote{Open-source simulation, deployment code, and videos are available at: \texttt{https://github.com/osheraz/IsaacGymInsertion}} for potential benchmarks and to advance research in the field. 
\end{enumerate}

%% file: related.tex
Precise robotic manipulations are often performed with rigid hands \cite{narang2022factory,Suh2022}. These hands are usually fully-actuated, making their state deterministic. Hence, having hand proprioception along with visual perception and tactile sensing may provide enough information for accurate pose estimation of a grasped object \cite{Bauza2024}. Nevertheless, rigid hands are often characterized by complex structures with numerous degrees of freedom 
\cite{Bai2014}. This complexity can lead to increased fragility, higher costs, and challenges in control. An alternative approach involves the use of compliant hands \cite{Dollar2010}. This strategy enhances flexibility and adaptability in grasping various objects, eliminating the need for precise closed-loop control. However, this same adaptability makes precise control more difficult due to the absence of accurate analytical models, inherent uncertainties, and no proprioception feedback \cite{Sintov2019}. Despite ongoing research, developing robust manipulation skills with compliant hands is still a key challenge \cite{weinberg2024survey}.

Combining extrinsic vision and tactile sensing is often termed \textit{Visuotactile} sensing and has been widely explored with rigid hands in contact-rich tasks such as insertion and in-hand manipulation \cite{Hansen2022, qi2023general}, often with known objects \cite{Bauza2024}. However, in rigid hands, the state of the hand is rather deterministic, and the visuotactile data sufficiently captures the state of the hand-object, exhibiting minimal uncertainty. Learning manipulation skills with compliant hands, on the other hand, presents significant challenges due to the uncertainties in the hand-object state \cite{morgan2021vision}.

Recently, learning-based methods have gained popularity in acquiring manipulation policies \cite{qi2023hand,chen2023visual}, offering the potential for greater generalization as more data becomes available. Azulay et al. \cite{azulay2022haptic} employed Force/Torque sensing at the wrist of a compliant hand for insertion tasks, requiring a substantial number of real-world training episodes. Notably, training with force sensing in simulation is often unreliable, making the transfer to a real-world robot infeasible. Recent progress in the area of learning-based methods stems from two main sources: the use of expert demonstrations combined with imitation learning \cite{Pari2022}, and learning in simulation paired with sim-to-real techniques \cite{Tobin2017,Gualtieri2018}. Both approaches attempt to compensate for the lack of diverse real-world data. However, these have limitations when it comes to manipulation with compliant hands. Expert demonstrations are typically acquired through teleoperation, which can be challenging when controlling compliant hands due to the lack of proprioceptive feedback and the inherent difficulties of remote manipulation.
Meanwhile, sim-to-real methods, while showing strong generalization and robustness across tasks like in-hand manipulation, grasping, and long-horizon manipulation, struggle to simulate the complex and uncertain dynamics of compliant hands \cite{sintov2020tools}.

%% file: method.tex
\subsection{Problem Formulation}

We consider a case of a robotic arm equipped with a three-finger compliant hand \cite{Dollar2010, Ma2017YaleOP}, where each fingertip of the hand includes a 3D round optical tactile sensor, as seen in Figure \ref{fig:intro}. Such a sensor was shown in previous work to be able to approximate the full contact state \cite{azulay2023allsight}. Furthermore, a single RGB-D camera is positioned near the robot to capture the object grasped by the hand. Our main approach involves decomposing raw RGB-D images into segmented point clouds focused on task-specific objects, creating object-centered 3D representations for policy learning. The camera can only acquire a partial point cloud derived from its point-of-view. The true state of the grasped object $\ve{s}\in SE(3)$ includes its position $\ve{x}=(x,y,z)$ and orientation $(\alpha,\gamma,\theta)$ relative to some coordinated frame, typically the base of the robot. As mentioned, joint angles are usually not available for compliant hands. Consequently, either the object's state is uncertain due to the initial grasp or due to involuntary perturbations originating from contact forces and hand compliance. Hence, explicit access to the true state of the object is not available. At time $t$, the policy can only acquire the End-Effector (EE) pose $\ve{x}_t\in SE(3)$, RGB-D image $\ve{I}_t\in\mathcal{I}$ and a set of three tactile images $\ve{T}_t\in\mathcal{J}^3$, where $\mathcal{I}$ and $\mathcal{J}$ are the spaces of depth and tactile images, respectively.

While the hand is grasping the object, the policy will attempt to insert the object into a nearby socket. In practice, the policy will take an action $\ve{a}_t\in SE(3)$ which is the change required for the hand pose $\ve{x}_t$ at time $t$. The following assumptions are considered:
\begin{enumerate}
    \item The object and socket match in shape with some clearance $\epsilon$, but the exact shape is not a priori known.
    \item The exact position and orientation of the socket are unknown, except for its visibility in the camera image.
    \item The object was previously grasped with some uncertainty in its $SE(3)$ pose.
    \item The state of the hand (e.g., finger positions and forces) is not known during the task.
\end{enumerate}
Due to the above assumptions, the target location of the hand for object-hole alignment is not known. The insertion goal is, therefore, to train a policy that can align the object and insert it into the socket with zero-shot deployment. 


\vspace{-2px}
\subsection{Visuotactile perception}

In this work, we train an insertion policy in simulation and then observe its performance with different perception modalities in the real world. EE pose is easily acquired in simulation and in the real system through simple forward kinematics of the robotic arm. However, visuotactile sensing requires careful adaptation for robust performance.

\subsubsection{Visual perception} Camera images are easily accessible in both simulation and the real world, while a significant gap exists when using color images. To address this, we solely use depth data acquired from the RGB-D camera. Inspired by the work in \cite{liu2024visual}, we use segmented depth images and masks to reduce background noise and highlight object-centric features. Specifically, we generate masks for both the object and the socket to ensure a more comprehensive representation. Although 2D segmentation masks can track objects over time, they are limited by specific camera viewpoints and may lack robustness against perspective changes. To achieve precise spatial reasoning, which is crucial for accurate insertion tasks, we extend the 2D masks by generating segmented point clouds for both the object and the socket. 
Point clouds $\ve{p}_{\text{object}} \in \mathbb{R}^{N \times 3}$ and $\ve{p}_{\text{socket}} \in \mathbb{R}^{M \times 3}$ are extracted from depth image $\ve{I}\in\mathcal{I}$ by back-projecting the 2D segmented masks into the 3D space using the known extrinsic parameters of the camera. This dual representation of the object and socket captures their geometry and enhances spatial reasoning, yielding improved insertion robustness.

\subsubsection{Tactile perception} 
\begin{figure}
\centering
\includegraphics[width=\linewidth,keepaspectratio]{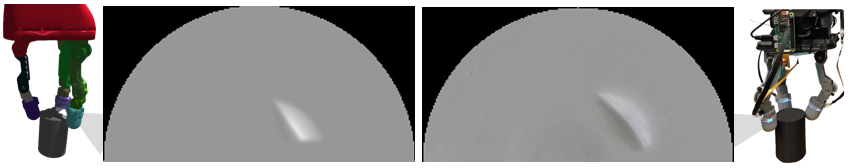} 
\vspace{-0.6cm}
\caption{\small Tactile images from (left) simulated and (right) real compliant hand with an AllSight sensor grasping similar objects.}
\label{fig:compare}
\vspace{-0.7cm}
\end{figure}

As mentioned, a tactile sensor is mounted at the tip of each of the three fingers. We consider high-resolution optical tactile sensors. Specifically, we integrate the AllSight optical sensor for its ability to provide implicit contact state approximation \cite{azulay2023allsight}. AllSight's round geometry is specifically designed for robust grasping in real-world scenarios. Nevertheless, other round sensors such as the 
RainbowSight \cite{Tippur} can be used with integration to the simulator. To simulate the sensor, we integrated TACTO \cite{wang2022tacto}, a rendering simulator for optical-based tactile sensors, into the IsaacGym simulator. 
The simulation was calibrated to sufficiently match the real-world by including reference images from real AllSight sensors \cite{azulay2023augmenting}, as demonstrated in Figure \ref{fig:compare}. Different reference images from several AllSight sensors were collected to enhance sim-to-real pre-training. Then, an image $\ve{T}_{ref}\in\mathcal{J}$ from the collected reference images is randomly picked and subtracted from the observed tactile image $\hve{T}_t\in\mathcal{J}$ yielding a reference image $\ve{T}_t=\hve{T}_i-\ve{T}_{ref}$. Such subtraction has been shown to enhance the learning process making the model agnostic to the background and focused only on color gradients that occur in the region of the contacts \cite{azulay2023allsight}. 


\subsection{Visuotactile Insertion Policy Learning}

 Directly training an insertion policy in the real world using Reinforcement Learning (RL) would be prohibitively data-intensive, requiring billions of samples for effective convergence. Hence, we train the insertion policy using the rigid body physics simulator IsaacGym for parallel training in a large set of environments, and then transfer the policy to the real system with no further training. Following prior work \cite{chen2023visual,liu2024visual, agarwal2023legged}, we adopt a two-phase training approach, leveraging privileged information in simulation and then distilling the knowledge into an object-centric representation that can be approximated using real-world sensing, as illustrated in Figure \ref{fig:sketch}. 

\subsubsection{Teacher Policy Training} During the initial phase, we learn a teacher policy using the state-of-the-art RL algorithm Proximal Policy Optimization (PPO). 
Since tactile and depth images are computationally expensive to render, and directly training RL on high dimensional observations is not always stable, we use privileged information about the environment to guide the RL policy to generate robust insertion actions in simulation. This privileged information includes task-relative information and physical properties. The task-relative features capture the robot's dynamics and spatial relationships with the environment, including hand and arm poses, velocities, as well as the positions and orientations of the socket and object, along with their relative errors. This results in a $50$-dimensional vector denoted as $\ve{e}_t^{task}$. The physical properties include the object mass, dimensions and coefficient of friction, resulting in a $13$-dimensional vector $\ve{e}_t^{phys}$. We define the complete privileged information at time $t$ by the concatenated vector $\ve{e}_t = [\ve{e}_t^{task}, \ve{e}_t^{phys}]$. Vector $\ve{e}_t$ is passed as an input to a privileged encoder 
which outputs an $n$-dimensional latent embedding $\ve{z}_t$. 

During RL training, a base policy $\pi_\theta$ takes the embedding $\ve{z}_t$ and an additional observation $\ve{o}_t$ to output an action $\ve{a}_t=\pi_\theta(\ve{o}_t,\ve{z}_t)$, where $\theta$ is the learnable parameters vector. For the model to be agnostic to the robot, observation $\ve{o}_t$ includes only the current EE pose $\ve{x}_{t}$ and the last action $\ve{a}_{t-1}$. We model the policy network as a Multi-Layer Perceptron (MLP). The base policy and the privileged encoder are jointly trained end-to-end to maximize the expected return $\mathbb{E}\{\sum_{t=0}^{T}\gamma^{t}r_t\}$, where $r_t$ is the reward at time $t$, $\gamma$ is the discount factor and $T$ is the maximum episode length.



The reward function $r_t$ is adopted from \cite{narang2022factory} and introduces additional auxiliary rewards to encourage smooth insertion behaviors. Specifically, the reward is given by
\begin{equation}
    \label{eq:loss}
    r_t = \lambda_{1}r_{d} + \lambda_{2}r_{e} + \lambda_{3}r_{o} + \lambda_{4}r_{a} + \lambda_{5}r_{w},
\end{equation}
where $\lambda_d,\lambda_e,\lambda_o,\lambda_a,\lambda_w>0$ are weight parameters. Reward component $r_{d} = -\|\ve{k}_o - \ve{k}_s\|$ penalizes the misalignment of four key-points distributed along the objects central axis $\ve{k}_o$ and the socket axis $\ve{k}_s$. Additionally, reward component $r_{e}>0$ is given if $\|\ve{x}_o - \ve{x}_s\| < \epsilon$ for some small $\epsilon$ and, $\ve{x}_o$ and $\ve{x}_s$ are the positions of the object and socket, respectively. This encourages engagement with the socket. Similarly, component $r_{o} = -\|\ve{q}_{o} - \ve{q}_{s}\|$ penalizes orientation misalignment relative to the socket, with $\ve{q}_{o}$ and $\ve{q}_{s}$ are quaternions for the object and socket orientations, respectively. Reward $r_{a} = -\|\ve{a}_t\|$ penalizes high value actions to stabilize motion, and reward $r_{w} = -\|\ve{a}_t - \ve{a}_{t-1}\|$ penalizes action differences for achieving smooth motion.

\subsubsection{Student Distillation} Once a teacher policy has been trained solely on privileged information, it is distilled to a student policy which receives visuotactile information. We employ a Transformer-based architecture 
to process the multimodal sensory input. Specifically, we stack the three tactile images $\ve{T}_{t}$, and process them through a set of convolutional layers to encode the tactile data into an $m$-dimensional representation $\ve{f}_{\text{tactile},t}$. 
For the visual observations, we first pass the point cloud through an initialized PointNet \cite{qi2017pointnet} to separately encode the segmented point clouds $\ve{p}_{\text{object},t}$ and $\ve{p}_{\text{socket},t}$, generating $l$-dimensional encodings $\ve{f}_{\text{object},t}$ and $\ve{f}_{\text{socket},t}$. These two encodings are then concatenated and passed through an MLP, producing a joint feature representation, $\ve{f}_{\text{joint},t}$. This joint encoding captures the spatial relationships between the object and the socket, which is crucial for insertion. We also create the observation encoding $\ve{f}_{\text{obs},t}$ by passing $\ve{o}_{t}$ through a separate MLP. As in the teacher policy, observation $\ve{o}_t$ includes the current EE pose $\ve{x}_t$ and last action $\ve{a}_{t-1}$.

The final feature vector at time $t$ is formed by concatenating the tactile encoding, the joint point cloud and the current observation encoding yielding $\ve{f}_t = [\ve{f}_{\text{tactile},t}, \ve{f}_{\text{joint}, t}, \ve{f}_{\text{obs},t}]$.  
A set of $k$ past feature vectors $\ve{f}_{t-k}, ..., \ve{f}_{t-1}, \ve{f}_{t}$ is then stacked and fed into the transformer. Standard 2D sinusoidal positional embeddings are applied to maintain 
consistency.

\begin{figure*}[t]
\centering
\includegraphics[width=0.85\textwidth]{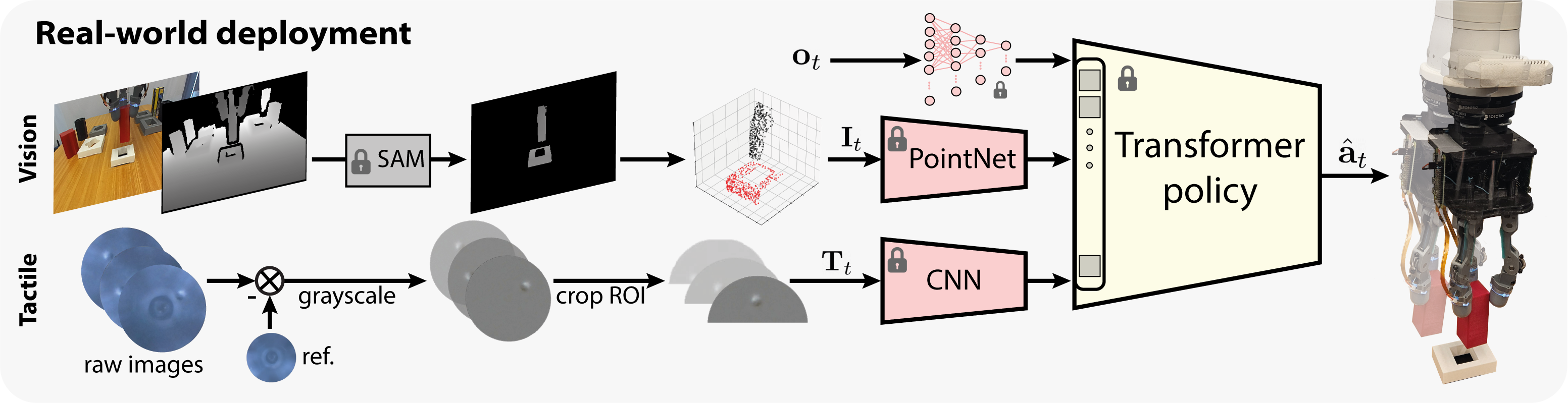} 
\vspace{-0.4cm}
\caption{\small A schematic illustrating the deployment of the student policy, trained in simulation, onto the real-world robot. The policy receives observation and visuotactile sensory data and generates control signals to actuate the robot arm.}
\label{fig:deploy}
\end{figure*}


The student policy is trained from scratch in two phases. In the first phase, we use Behavior Cloning (BC) to initialize the student policy. This involves collecting transition data by rolling out the pre-trained teacher policy in simulation, using privileged information to generate actions $\ve{a}_t$. Simultaneously, the student policy produces its own actions $\hat{\ve{a}}_t$ based on visuotactile sensing taken during the run of the teacher-guided simulation. In a supervised learning approach, the student policy is trained to minimize the $\ell_2$ distance between $\ve{a}_t$ and $\hat{\ve{a}}_t$ given the visuotactile input. This provides a strong prior for the policy. After the BC warm-up, we employ DAgger \cite{ross2011reduction} to refine the student policy. In this phase, the student policy generates actions, and the teacher provides corrective feedback as needed. Specifically, the teacher policy is used to collect trajectories with probability \( \beta \), while actions are sampled from the student with probability \( 1 - \beta \). Over the course of learning, \( \beta \) gradually decreases. 

The initial step of training a teacher policy with privileged inputs simplifies simulation training. However, the transition to real-world-compatible inputs introduces additional complexity during policy distillation. Hence, the two-stage process in the student training ensures that the final policy remains robust and deployable in real-world settings. As a result, robust and adaptive insertion behaviors emerge, leading to faster and more effective training compared to direct RL. 


\subsubsection{Bridging the sim-to-real gap}

Compliant hands, being inherently less stable than rigid hands, experience greater object variations during insertion attempts. To develop a robust insertion policy that can successfully transfer to real-world applications, we employ extensive domain randomization across various simulation parameters, including perception inputs, physical properties, and object attributes. These randomizations contribute to building a resilient low-level policy. To ensure the student policy remains effective and stable in both simulation and real-world environments, we have implemented several training techniques. 
First, a maximum of $10$-step action delay is introduced to account for inference and execution latencies that occur in real-world scenarios. With a small probability of 10\%, random perturbations to the socket pose are applied during training, helping the robot learn recovery behaviors when trials fail. Similarly, we apply random scaling to actions to accommodate varying relative EE pose motions. 

To enhance model robustness against visual and tactile variations, we implement a suite of perturbations during the training phase. Specifically, for the simulated tactile images, the acquired images were augmented by adding noise, varying the lighting conditions including gray-scaling and adding pixel-level Gaussian noise. For the visual observation, we add Gaussian noise to the camera pose and field of view. To simulate real-world segmentation imperfections, we use random pixel flips and simulated update delays. Similarly, random masking of up to 15\% of the object's visibility is applied to simulate partial observations. We also applied a set of augmentations to the point cloud data, including random point dropout to simulate sensor noise and occlusions, small jittering to simulate variations in real-world conditions, and random rotations and scaling to ensure invariance to object orientations and sizes during manipulation.

\subsection{Real world policy deployment} 
\label{sec:deploy}

The student policy is deployed to the real system with no further training. However, to ease the transfer and enable zero-shot run, the visuotactile information is processed in real time as illustrated in Figure \ref{fig:deploy}. For tactile information, the raw real-world tactile images are processed to match the simulated images by including gray-scaling and background subtraction, to maintain a nearly uniform background focusing on the foreground contact area. To ease the computation, only the half of the tactile image with the region of interest (ROI) is used. We found that this processing of the tactile images is quite effective in closing the gap between the simulation and the real-world.

For point cloud information, the process begins by annotating task-relevant objects using an interactive segmentation model, where instance segmentation masks are created with simple annotation on a single frame. Text prompts can also be used to automate this process, but this approach falls outside the scope of this work. We used TrackingSAM \cite{liu2024visual}, an annotate and real-time track model built on top Segment Anything (SAM) \cite{kirillov2023segment} at $10$Hz. These 2D masks are then back-projected into 3D point clouds using depth data, without relying on RGB information. The point clouds are transformed into the robot base frame invariant to camera perspectives. The sensed information is fed into the trained transformer policy to acquire the next action.


%% file: experiments.tex
This section presents the evaluation of the proposed approach. We begin by describing the simulation and hardware setups, followed by the results. Videos of the simulations and experiments can be seen in the supplementary material.


\subsection{Setup}

\textbf{Hardware details.} The setup, seen in Figure \ref{fig:intro}, includes the three-finger OpenHand Model-O compliant hand, where each finger features two compliant joints equipped with springs. Tendon wires run along the length of each finger and are connected to actuators. At the tips of the fingers, the AllSight tactile sensors are integrated. The compliant hand is mounted on a Kinova Gen3 arm. For visual sensing, we used the Zed Mini Stereo Camera. The entire system is managed using the Robot Operating System (ROS) on a local machine with a single RTX3060 GPU. Stream os the robot kinematics is available at a rate of $100$ Hz, with the tactile and image stream specifically provided at $30$ Hz. 

\textbf{Simulation details.} As previously mentioned, IsaacGym was used as the simulation platform to train the insertion policy. We simulated the real robotic system with a robotic arm, the three-finger OpenHand model-O compliant hand, an object sampled from our dataset, and a matching socket. Hand compliance was not included in the simulation; instead, position control was used with randomized parameters such as friction and finger mass. This approach ensures adaptability to a wide range of hand designs without requiring dependence on specific compliance characteristics.
These objects are designed with various prismatic shapes and clearances within the range $\epsilon \sim [0.125, 0.95]$~mm. To introduce variability, we randomize both the initial relative positions and orientations of the object and socket, setting $x, y, z \sim [-0.1, 0.1]$ meters. For the grasped object, we use a broad range of poses and orientations to account for high variation in its pose, simulating real-world scenarios with compliant hands. Specifically, we sampled the object’s angles to $\alpha, \gamma, \theta \sim [-20^\circ, 20^\circ]$, while the socket’s orientation is set to $\theta \sim [-5^\circ, 5^\circ]$, with both positioned in front of the robot. To further enhance the policy’s generalization, we vary the objects’ dimensions within the range $\eta \sim [0.95, 1.1]$, and, in accordance, the shape of the socket along with its height. These bounds were intentionally set to be tight to impose the learning of extreme low-clearance insertions. Although we initially sampled the object’s orientation from the range $\alpha, \gamma, \theta$, the orientation variability was further increased once the object was grasped. This additional variability reflects the unpredictable nature of real-world grasps with compliant hands, ensuring the policy is resilient to a wider range of orientation disturbances.

\textbf{Objects.} Six objects were used in the simulation for training: cylinder, small ellipsoid, triangular prism, trapezoidal prism, cuboid, and small hexagon prism. For evaluation on the real robot, 11 different object-socket pairs were used, including five of the objects used for training in the simulation and additional six novel objects not used in training: rectangular prism, large hexagonal prism, gum box, flower-shaped prism, heavy box (400 gram) and large ellipsoid prism. Each object-socket pair has unique physical properties. The dimensions and object-socket clearances are presented in Table \ref{tb:train_tab}. Also, the heights of the tested sockets were arbitrarily chosen as 10 mm and 20 mm.



\begin{figure}
\centering
\includegraphics[width=\linewidth]{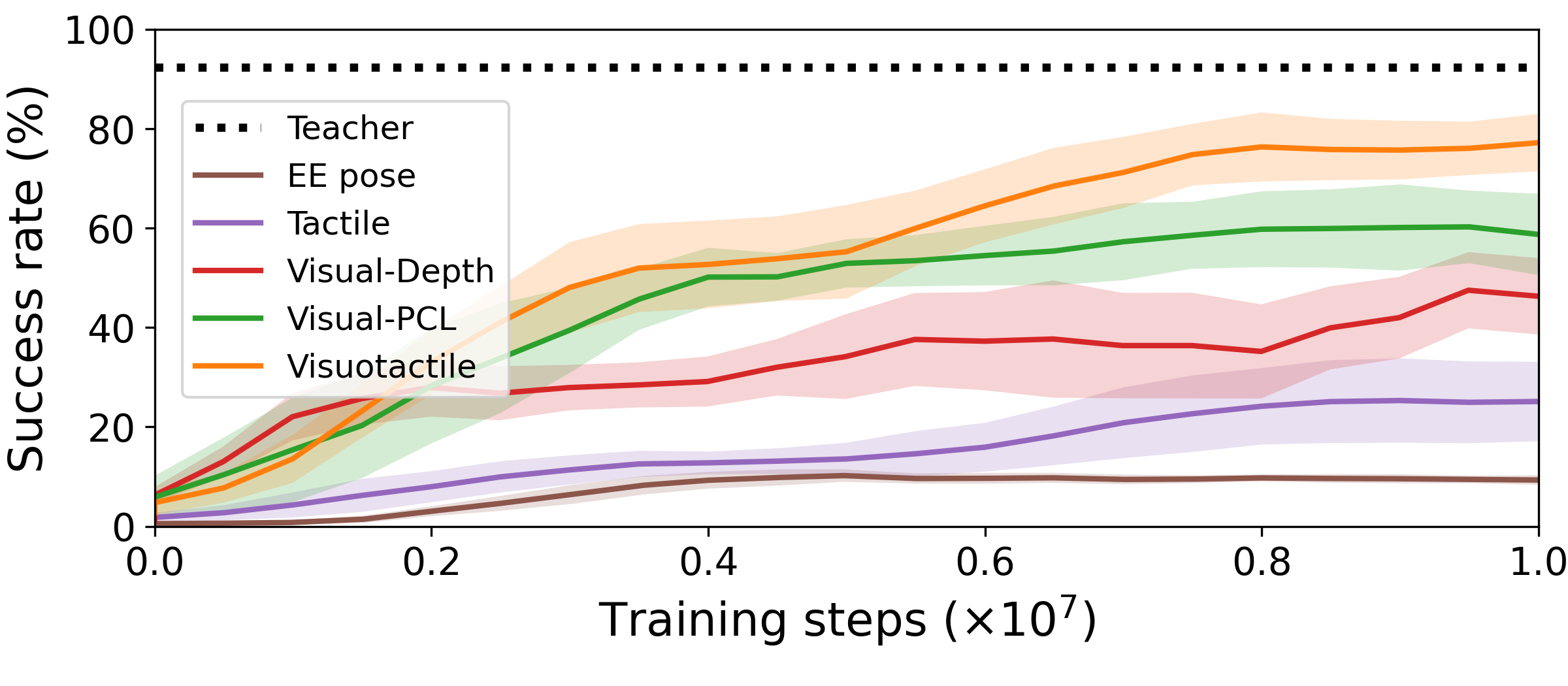}
\vspace{-0.8cm}
\caption{\small Ablation study of insertion success rate in simulation for policies trained with different sensory modalities, as a function of the number of training steps.}
\vspace{-0.4cm}
\label{fig:multi_experiment_plot}
\end{figure}
\begin{figure}
\centering
\includegraphics[width=\linewidth]{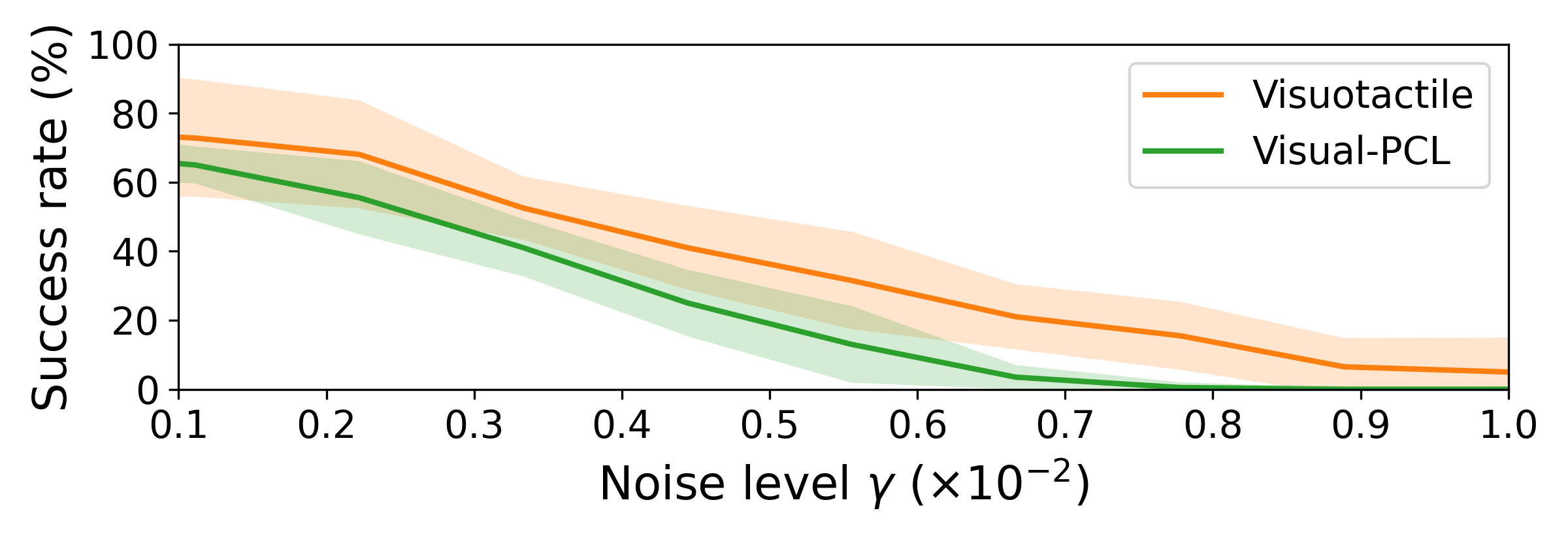} 
\vspace{-0.8cm}
\caption{\small Insertion success rate in simulation with regard to the noise level $\gamma$ added to the perceived point cloud.}
\vspace{-0.7cm}
\label{fig:aggregated_policy_comparison_plot}
\end{figure}
\begin{table*}
\caption{\small Insertion success rates of different policies with the real-world robot}
\label{tb:train_tab}
\centering
\begin{adjustbox}{width=\linewidth}
\begin{tabular}{lccccc c cccccccc}
\toprule
\multirow{2}{*}{Method} & \multicolumn{5}{c}{Train objects} & & \multicolumn{6}{c}{Test objects}\\
\cmidrule{2-6}\cmidrule{8-13}
 &
Cyl. &
Ellip. (S) &
Tri. &
Trapez.&
Hex. (S)&&
Rect.&
Hex. (L)&
Gum B. &
Flower &
Box &
Ellip. (L) \\ \midrule

 & 
\includegraphics[height=0.3in]{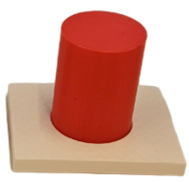} &
\includegraphics[height=0.3in]{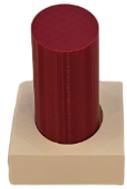} &
\includegraphics[height=0.3in]{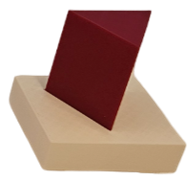} &
\includegraphics[height=0.3in]{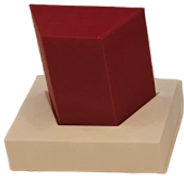} &
\includegraphics[height=0.3in]{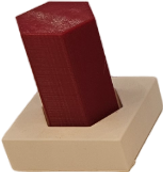} &&
\includegraphics[height=0.3in]{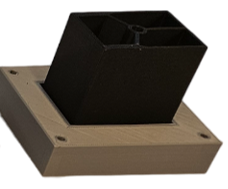} &
\includegraphics[height=0.3in]{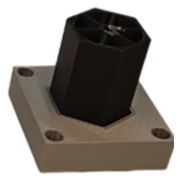} &
\includegraphics[height=0.3in]{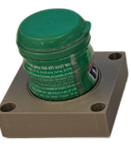} &
\includegraphics[height=0.3in]{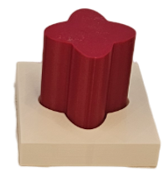} &
\includegraphics[height=0.3in]{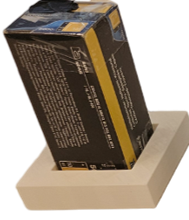} &
\includegraphics[height=0.3in]{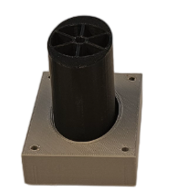} \\ 

Dimensions (mm) & 50 & $35\times50$ & $41\times47$ & $30\times60\times20$ & 40 & & $40\times60$ & 50 & 52 & $30\times60$ & $45\times45$ & $40\times60$ \\
Clearance $\epsilon$ (mm) & 0.6 & 1.0 & 0.7 & 1.2 & 1.0 && 1.2 & 1.1 & 0.35 & 1.2 & 2.0 & 1.35 \\\midrule



Visual-PCL & 80\% & 85\%  & 65\% & 50\% & \cellcolor[HTML]{C0C0C0}80\% && 55\% & 80\% & 75\% & 65\% & 60\% & 70\%  \\ 

Visuotactile w/o Diff. \& Aug. & 70\% & 60\% & 50\% & 40\% & 60\% && 50\% & 70\% & 70\% & 65\% & 70\% & 65\% \\
    
Visuotactile & \cellcolor[HTML]{C0C0C0}95\% & \cellcolor[HTML]{C0C0C0}90\% & \cellcolor[HTML]{C0C0C0}85\% & \cellcolor[HTML]{C0C0C0}65\% & \cellcolor[HTML]{C0C0C0}80\% && \cellcolor[HTML]{C0C0C0}75\% & \cellcolor[HTML]{C0C0C0}85\% & \cellcolor[HTML]{C0C0C0}90\% & \cellcolor[HTML]{C0C0C0}85\% & \cellcolor[HTML]{C0C0C0}90\% & \cellcolor[HTML]{C0C0C0}80\% \\  \bottomrule    
\end{tabular}
\end{adjustbox}
\end{table*}

\begin{figure*}
\centering
\includegraphics[width=0.9\linewidth]{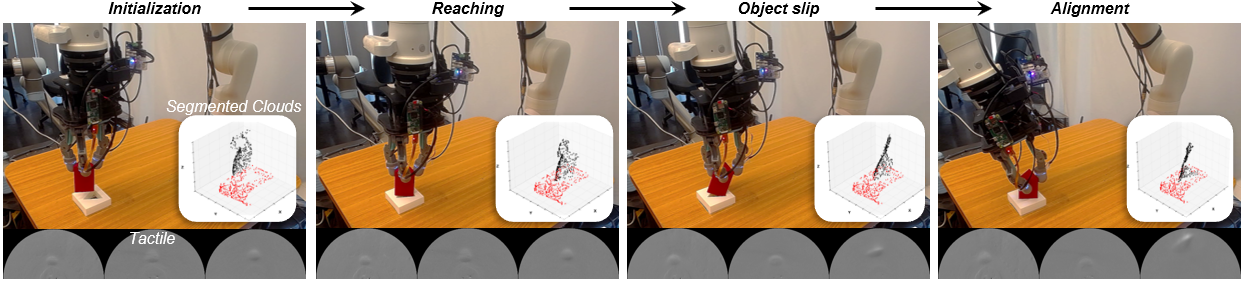} 
\vspace{-0.2cm}
\caption{\small A visuotactile policy deployment trial for inserting a triangular prism. The instantaneous views of the tactile images and point cloud are illustrated.} 
\label{fig:example}
\vspace{-0.7cm}
\end{figure*}

\subsection{Simulation Results}

For training the teacher policy, 4,096 parallel environments were deployed. However, the visuotactile student policy requires high GPU memory due to visual and tactile scene rendering, forcing the use of only 32 parallel environments. All trainings were conducted on a remote machine with a single RTX4090 GPU, with the teacher and student policies lasting approximately 24 and 48 hours, respectively. The training times of the teacher and student correspond to approximately $10^8$ and $\num{0.7e7}$ episodic states, respectively. For loss function \eqref{eq:loss}, a parameter sweep suggested the following parameter values $\lambda_{1}=-0.9$, $\lambda_{2}=0.4$, $\lambda_{3}=-0.5$, $\lambda_{4}=-0.1$ and $\lambda_{5}=-0.1$, yielding best results. We perform an ablation study and compare our visuotactile policy to several policies trained with different sensing modalities. The tested variations include: the \textit{Teacher} policy with privileged data; using only the \textit{EE pose} observations; using \textit{Tactile} sensing; \textit{Visual-Depth} perception based on the segmented depth \cite{liu2024visual} without projecting it into a point-cloud; visual observations based on segmented point-clouds (\textit{Visual-PCL}); and \textit{Visuotactile} sensing combining both tactile and visual perception as proposed in this work. The latter four variations are used alongside the EE pose observations. 
For a fair comparison, since the socket position varies, both the Tactile and EE pose policies also receive noisy estimates of the socket position in the observation $\ve{o}_t$.

In preliminary tests, we evaluated Rapid Motor Adaptation (RMA) \cite{kumar2021rma}, which distills the teacher policy using a latent space of environment extrinsics. While RMA performed well on a single object-socket pair, it struggled to generalize across multiple pairs, likely due to a realizability gap \cite{liu2024visual}. Furthermore, our findings indicate that asynchronous restarting and diversified data sampling are crucial for robust visuotactile policy, leading to a 50\% success rate improvement.


Figure \ref{fig:multi_experiment_plot} presents the learning curves for insertion success rate in simulation for each of the policies with regard to the training steps. Each policy was trained three times with different random seeds to acquire mean and standard deviation over the success rate. The baseline performance, achieved by a teacher policy with privileged information, yielded a maximum insertion success rate of 92.4\%. EE pose alone without any external perception proves insufficient for accurate hand-object and object-socket pose estimation, leading to poor performance. Adding tactile slightly improves performance but still performs ineffective. It captures only local contact states without broader visual context, making it difficult to align the object with the socket accurately due to a lack of hand proprioception. Policies with Visual-PCL and Visual-Depth, on the other hand, acquired higher success rates due to the additional contextual information provided by the visual data. Visual-PCL leverages segmented point clouds to understand the object's shape and orientation, offering strong performance by capturing structural information compared to Visual-Depth. Finally, the results clearly demonstrate the advantage of the visuotactile policy, which benefits from the combined information provided by tactile feedback and segmented point cloud data. This dual input provides detailed structural and contact information, allowing the policy to make precise adjustments during insertion.

 
To evaluate the importance of integrating tactile sensing with visual information, we explore the effect of adding noise to the visual observation on the insertion policy success rate. Figure \ref{fig:aggregated_policy_comparison_plot} presents the success rate for both the Visuotactile and the Visual-PCL policies with respect to noise level $\gamma$ added to the point cloud. The noise level is the variance of the Gaussian noise $\mathcal{N}(0,\gamma)$ added to the point cloud. For each level, 100 trial runs were averaged. Similar to the findings in \cite{suresh2023neural}, our results show that the integration of tactile sensing enhances accuracy, particularly in high-noise scenarios.



\subsection{Real-World Results}

We next transfer the student policies trained entirely in simulation to the real system as described in Section \ref{sec:deploy}, and evaluate the zero-shot performances. We compare between Visual-PCL and Visuotactile policies. In addition, we evaluate a Visuotactile policy without tactile image differences and image augmentation in simulation. The policies are evaluated on 11 objects, six of which are novel and not encountered during simulation training. Table \ref{tb:train_tab} reports the zero-shot insertion success rate of the different policies along with dimensions and clearances of the tested objects. For each object and policy, 20 trials were deployed where a trial is completed if the object successfully slides into the socket or fails after 1,000 steps. Here also, the results show that using a segmented point cloud with Visual-PCL achieves strong accuracy, while the addition of tactile information further boosts performance significantly. Additionally, the Visuotactile policy without difference and augmentation performs poorly. This emphasizes the importance of using difference tactile images and comprehensive image augmentation in simulation to bridge the sim-to-real gap for tactile data. Our results demonstrate strong generalization to novel objects, achieving performance comparable to that on training objects, highlighting the robustness of the approach. Failures in insertion tasks are generally attributed to unstable grasps, leading to object slippage and potential drops. Figure \ref{fig:example} shows a demonstration of inserting a triangular prism. Similarly, Figure \ref{fig:perturbations} demonstrates a successful insertion attempt despite manual perturbations to the object. The robot's ability to recover and complete the task, even in the presence of disturbances like object perturbations, socket pose variation and temporary occlusions (as shown in the supplementary video), highlights its robust insertion capabilities.


%% file: Conclusions.tex
In this work, we presented a framework for learning a robust precision insertion policy for low-cost compliant hands. We introduced a two-stage teacher-student architecture fully trained in simulation and then deployed on real hardware with zero-shot transfer. First, a teacher policy with access to privileged information is trained using RL, which is subsequently distilled into a visuotactile policy through imitation learning. Our system is extensively tested in both simulation and real-world settings, demonstrating robust, generalized performance across various object-socket pairs. This approach is adaptable to different hands and environments. While visual perception can readily adapt to new environments, a limitation arises from the need for the simulation to accurately model the selected tactile sensor, requiring potential adaptation.

\begin{figure}
\centering
\includegraphics[width=\linewidth]{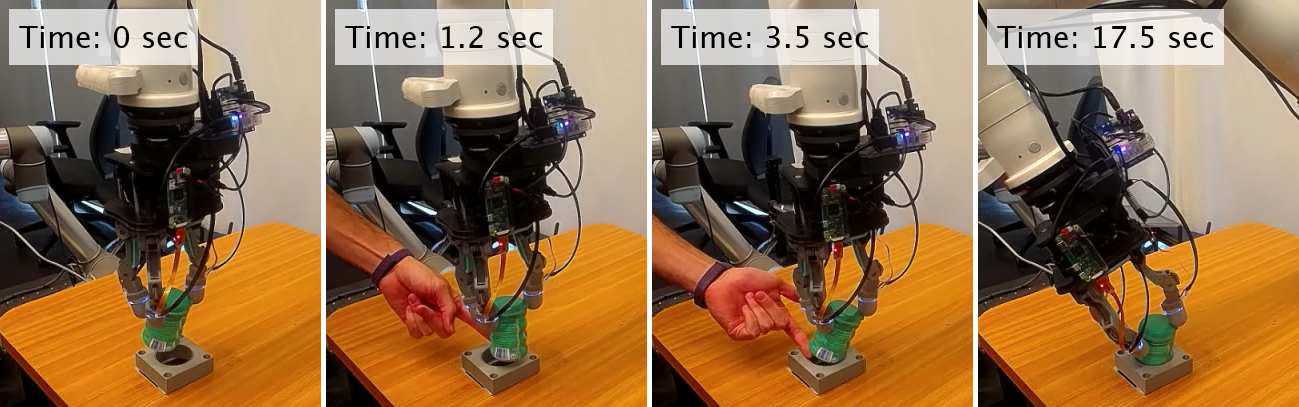} 
\vspace{-0.6cm}
\caption{\small Demonstration of manual interferences to the deployment of the insertion policy. The robot is able to overcome the interferences and successfully insert the object.}
\vspace{-0.6cm}
\label{fig:perturbations}
\end{figure}

By examining the role of each modality, we emphasize the importance of visual task-specific perception in enhancing tactile sensing for adaptable and successful insertion. Designing effective policies for compliant hand manipulation requires representations that are object-focused, spatially aware, and sensitive to relative positioning within visuotactile interactions. These qualities support task-relevant focus, 3D spatial reasoning, and subtle contact detection. 
While this work focuses on insertion as a test case, the methodology inherently supports adaptability to variations within this task by modifying simulation parameters and environmental conditions. Nevertheless, the approach relies on task-specific tuning which needs to be addressed in new tasks. Hence, future work could explore extending this framework to more complex manipulation tasks, such as dexterous object reorientation, in-hand manipulation or multi-step assembly, where longer control planning and adaptability are further challenged by object dynamics and multiple object-contact interactions. Future work could also explore leveraging depth alongside RGB data to enhance spatial reasoning and address challenges in reflective and transparent surfaces.







